\UseRawInputEncoding

\documentclass[UTF8,a4paper,conference]{IEEEtran}

\usepackage{multicol}

\begin{multicols}{1}
\end{multicols}

\usepackage{graphicx}
\usepackage{amsthm}
\usepackage{subfigure}
\usepackage{amssymb, amsmath, amsfonts, epsfig, latexsym, times}
\usepackage{bm}
\usepackage{cite}
\usepackage{flushend}
\usepackage{booktabs}
\usepackage{multirow}

\usepackage{booktabs} 

\usepackage{algorithmic}
\usepackage{algorithm}

\usepackage{fancyhdr}

\begin{document}
%

\title{Efficient Adaptive Federated Optimization of Federated Learning for IoT}

\author{\IEEEauthorblockN{Zunming Chen$^{1}$, Hongyan Cui$^{1}$, Ensen Wu$^{1}$, Yu Xi$^{1}$}
\IEEEauthorblockA{1. State Key Lab. of Networking and Switching Technology, Beijing University of Posts and Telecommunications, 100876, China}
}
\IEEEpubid{978-1-5090-0690-8/16/$31.00 ?\copyright ?2016 IEEE}
\maketitle

\thispagestyle{fancy}
\fancyhead{}
\lhead{}
\lfoot{}
\cfoot{}
\rfoot{}

\begin{abstract}


The proliferation of the Internet of Things (IoT) and widespread use of devices with sensing, computing, and communication capabilities have motivated intelligent applications empowered by artificial intelligence. The classical artificial intelligence algorithms require centralized data collection and processing which are challenging in realistic intelligent IoT applications due to growing data privacy concerns and distributed datasets. Federated Learning (FL) has emerged as a distributed privacy-preserving learning framework that enables IoT devices to train global model through sharing model parameters. However, inefficiency due to frequent parameters transmissions significantly reduce FL performance. Existing acceleration algorithms consist of two main type including local update considering trade-offs between communication and computation and parameter compression considering trade-offs between communication and precision. Jointly considering these two trade-offs and adaptively balancing their impacts on convergence have remained unresolved. To solve the problem, this paper proposes a novel efficient adaptive federated optimization (EAFO) algorithm to improve efficiency of FL, which minimizes the learning error via jointly considering two variables including local update and parameter compression and enables FL to adaptively adjust the two variables and balance trade-offs among computation, communication and precision. The experiment results illustrate that comparing with state-of-the-art algorithms, the proposed EAFO can achieve higher accuracies faster.

\end{abstract}


\IEEEpeerreviewmaketitle

\section{Introduction}


The explosive growth of the amount of data from devices has witnessed the rapid development of the Internet of Things that provides ubiquitous sensing, computing and communication capabilities to connect things to the Internet \cite{1st}. To provide deep analysis for data from IoT devices, artificial intelligence (AI) algorithms have been adopted to enable intelligent IoT applications such as smart transportation and smart city \cite{2nd}. Traditionally, the AI algorithms that require centralized data collection and processing are deployed on a centralized cloud/edge server or a data center for data mining. However, the offloading of massive IoT data to the remote server and the processing of data in the remote server induce significant delays. Furthermore, the third-party server also raises data privacy concerns \cite{3rd}. In this context, integrating privacy-preservation and distributed AI into IoT becomes an important topic.

Federated Learning (FL) has recently emerged as a distributed privacy-preserving learning framework that enables intelligent IoT applications via allowing distributed IoT devices to train machine learning models collaboratively through sharing local model parameters instead of raw data \cite{4th}. FL enables multiple devices to train a joint global model via Stochastic Gradient Descent (SGD) and sharing local model parameters \cite{5th}. Although only model parameters rather than raw data are shared in FL, FL-IoT implementation still suffers from the inefficiency and limited convergence performance due to three reasons: 1) training data at devices are different in size and distribution because of different sensing environments, Thus, communications in uplinks and downlinks in FL are highly sensitive subject to the unbalanced and non-IID data. 2) direct and frequent transmissions of model parameters from devices to server will significantly reduce the communication performance of federated learning when the number of devices grows exponentially. 3) the convergence of FL for IoT can not be guarantee all the time due to the intermittent connections and heterogeneous computing and communication capabilities of devices \cite{6th,7th,8th}. The situation of federated learning on mobile devices (e.g., sensors and Unmanned Aerial Vehicles (UAVs)) gets even worse, which communicates via wireless channels and suffers from lower bandwidth, higher latency and intermittent connections \cite{9th}. Thus, the inefficiency and limited convergence performance problem becomes the important bottleneck for scaling up federated learning.

It is necessary to solve the inefficiency and limited convergence performance problem. Several works have been done to speedup federated learning convergence via local update and parameter compression \cite{10th,11th}. The approach of local update aims to reduce the frequent transmissions of model parameters via making full use of the computing capability of devices. The local update algorithms characterize trade-off between computation and communication by the proposed concept of local model update coefficient which determines the ratio of the number of local update to bandwidth \cite{12th}. The approach of parameter compression aims at reducing the amount of data to be transmitted through data compression techniques such as quantization and sparsification. The parameter compression algorithms characterize trade-off between communication and model precision via the parameter compression coefficient (compression budget) that determines the ratio between bandwidth and accuracy \cite{13th}. These methods have been individually studied to improve efficiency of federated learning. However, jointly considering these two trade-offs and adaptively balancing their impacts on convergence of federated learning from both mathematical estimation and theoretical analysis perspectives have remained unresolved. This significant problem motivated our research. This paper proposes a novel efficient adaptive federated optimization (EAFO) algorithm to speedup convergence of federated learning for IoT, which minimizes the learning error via jointly considering two methods including local update and parameter compression and enables federated learning to adaptively adjust the two variables and balance trade-offs among computation, communication and accuracy.

EAFO differs from current existing works in jointly considering two variables including local update coefficient characterizing trade-off between computation and communication and compression budget characterizing trade-off between communication and precision. The key contributions of the EAFO are summarized as the following:
\begin{itemize}
\item[$\bullet$]We investigate the federated learning problem with a practical formulation of minimizing the error of global model in terms of local update coefficient and compression budget, which characterizes trade-off between computation and communication and trade-off between communication and precision respectively.
\end{itemize}
\begin{itemize}
\item[$\bullet$]We propose a novel efficient adaptive federated optimization algorithm using a derived upper bound of error of global model considering two variables including local update and compression budget, which adaptively adjusts these two variables to improve efficiency of federated learning.
\end{itemize}
\begin{itemize}
\item[$\bullet$]Besides theoretical analysis of the proposed algorithm, we demonstrate strong empirical performance on three datasets of the proposed EAFO comparing with other state-of-the-art algorithms, which achieve higher accuracies faster.
\end{itemize}

The rest of this paper is organized as the followings. Section II discusses the related work about approaches to speedup federated learning including multiple local update and parameter compression. Section III provides preliminaries and mathematical problem formulation. Section IV introduces the technical details of EAFO, a novel efficient adaptive federated optimization algorithm. Section V provides performance evaluation of EAFO. Finally, Section VI concludes this paper with future research directions.

\section{Related Work}
\subsection{Multiple local update in federated learning}
The approaches to improve efficiency of federated learning and overcome communication bottleneck can be categorized into multiple local update and parameter compression. The approach of local update aims at taking full advantage of the computing capability of devices to reduce the frequent transmissions of model parameters. For example, Nenghai et al. have proposed ASGD which derives the bound of convergence of distributed gradient descent and only allows one step of local update before global aggregation \cite{14th}. Virginia et al. have proposed Fedprox that puts forward the concept of local model update coefficient determining the ratio of computation to communication, and it performs multiple local update with a fixed local update coefficient \cite{15th}. Joshi et al. have proposed ADACOMM, where an adaptive communication strategies is adopted to adjust local update coefficient and balance the trade-off between communication and computation dynamically to solve the problem of heterogeneous computing and communication capabilities of devices \cite{16th}. Kevin et.al have proposed AFD that adopts an adaptive strategies to determine the best local update coefficient under a given resource budget in order to speed up federated learning in both both iid and non-iid settings \cite{17th}.

\subsection{Parameter compression in federated learning}
The approaches of parameter compression uses data compression techniques such as quantization and sparsification to significantly reduce the amount of data to be transmitted. Parameter compression algorithm characterizes trade-off between communication and precision by compression coefficient determining the ratio of communication to precision. The approach of parameter compression reduce communication overheads via uploading the quantized version or sparse representation of the model parameters. For instance, Wen et al. have proposed Terngrad which quantizes each parameter to 2 bits \cite{18th}. Alistarh et al. have proposed QSGD which uses 2 bits, 4 bits, and 8 bits to quantize different layers of neural networks \cite{19th}. Seide et al. have proposed 1-bit SGD which even quantizes the parameter to one bit \cite{20th}. The sparsification techniques aims to sparsify parameters to send the significant parameters rather than all parameters. The sparsification techniques can be categorized into two categories according to the domain the sparsity is sought for, which includes raw domain and transformed domain. For example, Dally et al. have proposed a parameter sparsification strategy to set unimportant elements to zero via a threshold which defines values of unimportant elements are between top $0.05\%$ and bottom $0.05\%$ \cite{21th}. Wright et al. have proposed ATOMO which transforms parameters to domain of Singular Value Decomposition (SVD) to exploit the low-dimensional structure to obtain more sparsity aiming at reducing the communication overheads \cite{22th}.

The aforementioned methods have been individually studied to overcome communication bottleneck and improve efficiency of federated learning. It is expected that integrating the two approaches would be more effective in speeding up federated learning. However, integrating the two approaches and jointly considering the trade-off between communication and computation and trade-off between communication and precision to adaptively balancing and adjusting their impacts on convergence of federated learning have remained unresolved. Thus, there is an urgent necessity to design a novel efficient adaptive federated optimization algorithm via jointly balancing trade-off between communication, computation and precision to supplement existing approaches.


\section{Preliminaries and Problem Formulation}
In order to further describe the impact of multiple local update and parameter compression, this paper first presents mathematical analysis on how coefficient of multiple local update and parameter compression affect the federated learning. Additionally, we theoretically formulate our problem and derive the error upper bound of federated learning, which jointly considers two variables including multiple local update and parameter compression.

\subsection{Federated learning}
Without losing generality, the analysis of federated learning is on the FedAvg which is the most widely adopted frameworks of federated learning \cite{22th}. The learning objective of the FedAvg can be given by:

\begin{equation}
\begin{aligned}
   w = {\min _w}\left\{ {F(w) \buildrel \Delta \over = \sum\limits_{n = 1}^N {{p^n}} {F^n}(w)} \right\}
\end{aligned}
\end{equation}

where {w} represents global model weights, $N$ represents the number of devices, $p^n$ corresponds to weight of the n-th device, $\sum\nolimits_{n = 1}^N {{p^n}}  = 1$, ${p^n} \in [0,1]$ and $F^n$ corresponds to local objective of the n-th device. Eqn.1 can be optimized by iterative exchange of model parameter between devices and the central server. Particularly, at the t-th round, local model of n-th device $w_{t,i}^n$ can be denoted as:

\begin{equation}
\begin{aligned}
   w_{t,i + 1}^n \leftarrow w_{t,i}^n - {\eta _{t,i}}\nabla {F^n}\left( {w_{t,i}^n,\varphi _{t,i}^n} \right)
\end{aligned}
\end{equation}

where ${\varphi _{t,i}^n}$ corresponds to the data samples of randomly selected mini-batch from the n-th device, $\eta _{t,i}$ represents the learning rate of the n-th device, i represents the i-th local updates. As soon as the n-th device performs $I_t$ number of consecutive local updates based on Eqn. 2, the aggregated weights of local model $\ell (w_t^n)$ due to multiple local updates can be obtained by:

\begin{equation}
\begin{aligned}
    \ell (w_t^n) \leftarrow \sum\limits_{i = 1}^{{I_t}} \ell  \left( {w_{t,i}^n,\varphi _{t,i}^n} \right)
\end{aligned}
\end{equation}

where $\ell \left( {w_{t,i}^n,\varphi _{t,i}^n} \right) = \nabla {F^n}\left( {w_{t,i}^n,\varphi _{t,i}^n} \right)$. $I_t$ represents the number of consecutive local updates. As the local update coefficient, $I_t$ determines the ratio of computation to communication and can be adjusted to balance trade-off between computation and communication.

The devices then compress their locally aggregated weights $\ell \left( {w_t^n} \right)$ to $\widehat \ell \left( {w_t^n} \right)$ with the sparsity budget denoted as ${\varepsilon _t}$ of parameter compression. The compressed locally aggregated weights are sent to the central server by devices \cite{23th,24th}. As the parameter compression coefficient, ${\varepsilon _t}$ can be adjusted to balance trade-off between communication and precision. The server aggregates all compressed locally aggregated weights from devices to obtain compressed global weights denoted as $\hat \ell \left( {{w_t}} \right)$, which can be described as:

\begin{equation}
\begin{aligned}
    \widehat \ell \left( {{w_t}} \right) = \frac{1}{N}\sum\limits_{n = 1}^N {\widehat \ell } \left( {w_t^n} \right)
\end{aligned}
\end{equation}

The latest global weights ${w_{t + 1}}$ can be updated via SGD as the followings:

\begin{equation}
\begin{aligned}
    {w_{t + 1}} \leftarrow {w_t} - \eta \widehat \ell \left( {{w_t}} \right)
\end{aligned}
\end{equation}

The central server sends the latest global weights ${w_{t + 1}}$ to devices involving in federated learning. The local model of devices can be updated via the received latest global weights. This is the complete process of federated learning with joint consideration of both multiple local update and parameter compression.

\subsection{Local update coefficient $I_t$ and parameter compression coefficient ${\varepsilon _t}$}
As mentioned before, local update coefficient $I_t$ characterizes trade-off between computation and communication. It is significant to balance the trade-off because one extreme case with low communication and high computation slows the federated learning convergence while reverse extreme case reduces the accuracy thereby decelerates federated learning. The parameter compression coefficient ${\varepsilon _t}$ characterizes trade-off between communication and precision where the balance of the trade-off is crucial due to one extreme case with high compression and low precision impedes aggregation of federated learning but contrary extreme case results in high communication overheads.

Before formulating our problem, we first define two variables including local update coefficient denoted as $I_t$ and parameter compression coefficient denoted as ${\varepsilon _t}$. $I_t$ corresponds to the number of local update, which determines the ratio of computation of communication. Instead of adopting fixed $I_t$ over iteration rounds, we dynamically adjust the $I_t$ during federated learning to reduce unnecessary transmissions of parameters. ${\varepsilon _t}$ corresponds to the sparsity budget of parameter compression, which determines the ratio of communication of precision. Rather than transmitting original parameters, the approach of parameter compression only sends approximation parameters which is the sparse representation of the parameters consisting of ${\varepsilon _t}$ number of basic components \cite{25th}. In practice, the basic components are usually negligible compared with original parameters.

\subsection{Problem formulation}
In the proposed EAFO algorithm, two variables including $I_t$ and ${\varepsilon _t}$ are jointly considered and adaptive adjusted to speed up federated learning. Theoretically, the problem is to find the optimal solution of $I_t$ and ${\varepsilon _t}$ at various communication rounds which minimizes the error of federated learning in a given time. The problem can be mathematically formulated as the followings:

\begin{equation}
\begin{aligned}
& {{{\min }_{\{ {I_t}\} ,\{ {\varepsilon _t}\} }}{E_{\{ \varphi _{t,i}^n\} }}\left[ {\min F\left( {{w_t}} \right)} \right]}\\
& {{\rm{ s}}{\rm{.t}}{\rm{.  }}\sum\limits_{t = 1}^{U_t} {\left( {{T_{comp}} + {T_{comm}}} \right)}  = T}
\end{aligned}
\end{equation}

where ${F\left( {{w_t}} \right)}$ corresponds to the global learning objective defined in Eqn.1 with the global weights ${{w_t}}$ at the t-th round. T represents the given time constraint during which the error is require to be minimized to avoid trivial solutions about solutions need too much time with big local update coefficient and parameter compression coefficient. In order to solve the problem in Eqn.6, the key is to derive expression of the error of federated learning describing the interdependence of $I_t$ and ${\varepsilon _t}$. However, such expression is almost impossible to obtain. Additionally, it is also hard to find the error upper bound which jointly considers local update and parameter compression of federated learning.

\subsection{Learning error upper bound}
To derive the error upper bound of federated learning, which jointly considers two variables including multiple local update and parameter compression, this subsection first derive the error upper bound in terms of $I_t$. Additionally, we derive the error upper bound with consideration of multiple local update described by $I_t$ and parameter compression described by ${\varepsilon _t}$.

We first make the following Assumptions 1-3 to present the analysis of error upper bound without parameter compression inspired by \cite{26th}.

\textbf{Assumption 1}. \emph{The global loss function is differentiable, and L-smooth: $||\nabla F({\bf{V}}) - \nabla F({\bf{W}}){\rm{||}} \le L||{\bf{V}} - {\bf{W}}||$ and there is a lower bound ${{\rm{F}}_{\inf }}$}.

\textbf{Assumption 2}. \emph{The variance of the global weights in mini-batch is bounded by: ${E_{\left\{ {\varphi _{t,i}^t} \right\}}}\left[ {{{\left\| {\ell \left( {{w_{_t}}} \right) - \nabla F\left( {{w_t}} \right)} \right\|}^2}} \right] \le \lambda {\left\| {\nabla F\left( {{w_n}} \right)} \right\|^2} + \delta $. $\delta $ corresponds to the variance between ${\ell \left( {{w_n}} \right)}$, $\lambda $ and $\delta $ are constants inversely proportional to mini-batch size }.

\textbf{Assumption 3}. \emph{The SGD is an unbiased estimator of the FGD: ${E_{\left\{ {\varphi _{t,i}^n} \right\}}}\left[ {\ell \left( {{w_t}} \right)} \right] = \nabla F\left( {{w_t}} \right)$ }.

Thus, the theorem for error upper bound without parameter compression can be denoted as follows.

\textbf{Theorem 1}. \emph{Let $T_{comp}$, $T_{comm}$, $L$ and $\delta$ be defined therein and Assumptions 1-3 hold. Choose the learning rate which satisfies $\eta L + {\eta ^2}{L^2}{I_t}\left( {{I_t} - 1} \right) \le 1$  }. Thus the learning error after n rounds within given time T in Eqn.6 is bounded by:

\begin{equation}
\begin{aligned}
\frac{{2\left[ {F\left( {{w_t}} \right) - {F_{{\rm{inf}}}}} \right]}}{{\eta T}}\left( {{T_{comp}} + \frac{{{T_{comm}}}}{{{I_t}}}} \right) + \frac{{\eta L\delta }}{N} + {\eta ^2}{L^2}\delta \left( {{I_t} - 1} \right)
\end{aligned}
\end{equation}

\emph{Proof.} See Appendix in \cite{8th}.

Theorem 1 describes the trade-off between computation and communication to minimize learning error. The local update coefficient $I_t$ determines the ratio of computation to communication. Eqn.7 illustrates that the $I_t$ is in the numerator and the denominator of expression of error upper bound, which means error upper bound will decrease either the values of $I_t$ is too small or too large. Thus, it is necessary to achieve the balance. Additionally, the trade-off need to be dynamically and adaptively adjusted over various rounds of federated learning due to the dynamically varying loss function ${F\left( {{w_k}} \right)}$ is in Eqn.7.

Apart from local update, the approach of parameter compression which introduces compression into locally aggregated weights will complicate the analysis of error upper bound in two aspects: (i) $T_{comm}$ will be affected by parameter compression coefficient ${\varepsilon _t}$. (ii) The variance denoted as $\delta $ in Eqn.7 will depend on the parameter compression coefficient ${\varepsilon _t}$.

As the mentioned before, the compressed parameters is approximated by basic components in parameter compression. Because of sparsity, several components is more significant than others in the aspects of approximating the raw parameters \cite{27th,28th}. Thus, the problem is to select basic components unbiasedly to minimizes the variance $\delta $. The locally aggregated weights from n-th device can be rewritten as:

\begin{equation}
\begin{aligned}
\ell \left( {w_t^n} \right) = \sum\limits_{k = 1}^K {{{d^k}}} \left( {w_t^n} \right){\alpha ^k}\left( {w_t^n} \right)
\end{aligned}
\end{equation}

where K is the number of basic components, ${\alpha ^k}\left( {w_n^t} \right)$ corresponds to the k-th basic component and ${d ^k}\left( {w_t^n} \right)$ is the corresponding weight. Our analysis is based on the fact a matrix can be denoted as a combination of basic matrices, which is the atomic decomposition for sparse representation in compressed sensing. Thus, our analysis can be extended nearly all unbiased compression. For example, TernGrad \cite{18th} and QSGD \cite{19th} are special cases of the Eqn.8. Additionally, sparsification algorithm including ATOMO \cite{22th} also follow the Eqn.8. The formulation presents a problem on selection of ${d ^k}\left( {w_t^n} \right)$. To meet the requirement of unbiased selection, we propose a estimator as follows inspired by \cite{22th}:

\begin{equation}
\begin{aligned}
\hat \ell \left( {w_t^n} \right) = \sum\limits_{k = 1}^K {\frac{{{d ^k}\left( {w_t^n} \right){e^k}\left( {w_t^n} \right)}}{{{p^k}\left( {w_t^n} \right)}}} {\alpha ^k}\left( {w_t^n} \right)
\end{aligned}
\end{equation}

where ${{p^k}\left( {w_t^n} \right)}$ corresponds to the probability characterizing the Bernoulli distribution, ${p^k}\left( {w_t^n} \right) \in (0,1]$ and ${{e^k}\left( {w_t^n} \right)}$ obeys Bernoulli distribution. We provide two significant properties for the estimator via the Lemma 1 and Lemma 2.

\textbf{Lemma 1} The variance of estimator given by Eqn.9 can be denoted as: ${E_{\left\{ {{e^k}\left( {w_t^n} \right)} \right\}}}\left[ {{{\left\| {\widehat \ell \left( {w_t^n} \right) - \ell \left( {w_t^n} \right)} \right\|}^2}} \right] = \sum\limits_{k = 1}^K {{d ^k}} {\left( {w_t^n} \right)^2}\left( {\frac{1}{{{p^k}\left( {w_t^n} \right)}} - 1} \right)$.

\textbf{Lemma 2} The estimator in Eqn.9 is unbiased, which means: ${E_{\left\{ {{e^k}\left( {w_t^n} \right)} \right\}}}\left[ {\widehat \ell \left( {w_t^n} \right)} \right] = \ell \left( {w_t^n} \right)$.

\emph{Proof.}  The variance of the estimator in Eqn. 9 is defined as follows:

\begin{equation}
\begin{aligned}
& {E_{\left\{ {{e^k}\left( {w_t^n} \right)} \right\}}}\left[ {{{\left\| {\widehat \ell (w_t^n) - \ell (w_t^n)} \right\|}^2}} \right] \\
& { = {E_{\{ {e^k}\left( {w_t^n} \right)\} }}\left[ {||\sum\limits_{k = 1}^K {\left( {\frac{{{d^k}\left( {w_t^n} \right){e^k}\left( {w_t^n} \right)}}{{{p^k}\left( {w_t^n} \right)}}} \right)} {a^k}\left( {w_t^n} \right)} \right.} \\
& {\left. { - {d^k}\left( {w_t^n} \right){a^k}\left( {w_t^n} \right)|{|^2}} \right]} \\
& { = {E_{\left\{ {{e^k}\left( {w_t^n} \right)} \right\}}}\left[ {\left( {\sum\limits_{k = 1}^K {{d^k}\left( {w_t^n} \right){a^k}\left( {w_t^n} \right)} } \right.} \right.} \\
& {{{\left. {\left( {\frac{{{e^k}\left( {w_t^n} \right) - {p^k}\left( {w_t^n} \right)}}{{{p^k}\left( {w_t^n} \right)}}} \right)} \right)}^T} \times \left( {\sum\limits_{k = 1}^K {{d^k}} \left( {w_t^n} \right){a^k}\left( {w_t^n} \right)} \right.} \\
& {\left. {\left. {\left( {\frac{{{e^k}\left( {w_t^n} \right) - {p^k}\left( {w_t^n} \right)}}{{{p^k}\left( {w_t^n} \right)}}} \right)} \right)} \right]} \\
& { = \sum\limits_{k = 1}^K {{d^k}} {{\left( {w_t^n} \right)}^2}||{a^k}\left( {w_t^n} \right)|{|^2}} \\
& { \times {E_{\left\{ {{e^k}\left( {w_t^n} \right)} \right\}}}\left[ {{{\left( {\frac{{{e^k}\left( {w_t^n} \right) - {p^k}\left( {w_t^n} \right)}}{{{p^k}\left( {w_t^n} \right)}}} \right)}^2}} \right]} \\
& { + \sum\limits_{x,y;x \ne y}^k {{d^x}} \left( {w_t^n} \right){d^y}\left( {w_t^n} \right)\left\langle {{a^x}\left( {w_t^n} \right),{a^y}\left( {w_t^n} \right)} \right\rangle }\\
& { \times {E_{\left\{ {{e^k}\left( {w_t^n} \right)} \right\}}}\left[ {\left( {\frac{{{e^x}\left( {w_t^n} \right) - {p^x}\left( {w_t^n} \right)}}{{{p^x}\left( {w_t^n} \right)}}} \right)} \right.} \\
& {\left. { \times \left( {\frac{{{e^y}\left( {w_t^n} \right) - {p^y}\left( {w_t^n} \right)}}{{{p^y}\left( {w_t^n} \right)}}} \right)} \right].}
\end{aligned}
\end{equation}
where ${{E_{\left\{ {{e^k}\left( {w_t^n} \right)} \right\}}}\left[ {{{\left( {\frac{{{e^k}\left( {w_t^n} \right) - {p^k}\left( {w_t^n} \right)}}{{{p^k}\left( {w_t^n} \right)}}} \right)}^2}} \right]}$ can be denoted as:

\begin{equation}
\begin{aligned}
& {E_{\left\{ {{e^k}\left( {w_t^n} \right)} \right\}}}\left[ {{{\left( {\frac{{{e^k}\left( {w_t^n} \right) - {p^k}\left( {w_t^n} \right)}}{{{p^k}\left( {w_t^n} \right)}}} \right)}^2}} \right] \\
& = \left( {1 - {p^k}\left( {w_t^n} \right)} \right) \times {\left( {\frac{{0 - {p^k}\left( {w_t^n} \right)}}{{{p^k}\left( {w_t^n} \right)}}} \right)^2} \\
& + {p^k}\left( {w_t^n} \right) \times {\left( {\frac{{1 - {p^k}\left( {w_t^n} \right)}}{{{p^n}\left( {w_t^n} \right)}}} \right)^2} \\
& = \left( {\frac{1}{{{p^k}\left( {w_t^n} \right)}} - 1} \right)
\end{aligned}
\end{equation}
and ${E_{\left\{ {{e^k}\left( {w_t^n} \right)} \right\}}}\left[ {\left( {\frac{{{e^x}\left( {w_t^n} \right) - {p^x}\left( {w_t^n} \right)}}{{{p^x}\left( {w_t^n} \right)}}} \right)} \right]$ can be obtained by:

\begin{equation}
\begin{aligned}
& {E_{\left\{ {{e^k}\left( {w_t^n} \right)} \right\}}}\left[ {\left( {\frac{{{e^x}\left( {w_t^n} \right) - {p^x}\left( {w_t^n} \right)}}{{{p^x}\left( {w_t^n} \right)}}} \right)} \right] \\
&  = \left( {1 - {p^x}\left( {w_t^n} \right)} \right)\left( {\frac{{0 - {p^x}\left( {w_t^n} \right)}}{{{p^x}\left( {w_t^n} \right)}}} \right) \\
&+ {p^x}\left( {w_t^n} \right)\left( {\frac{{1 - {p^x}\left( {w_t^n} \right)}}{{{p^x}\left( {w_t^n} \right)}}} \right)= 0.
\end{aligned}
\end{equation}
and in the same way, ${E_{\left\{ {{e^k}\left( {w_t^n} \right)} \right\}}}\left[ {\left( {\frac{{{e^y}\left( {w_t^n} \right) - {p^y}\left( {w_t^n} \right)}}{{{p^y}\left( {w_t^n} \right)}}} \right)} \right] = 0$.

Thus, based on Eqn. 10, 11, 12 and ${\left\| {{a^k}\left( {w_t^n} \right)} \right\|^2} = 1$ the variance of the estimator in Eqn. 9 can be obtained as follows:
\begin{equation}
\begin{aligned}
& {E_{\left\{ {{e^k}\left( {w_t^n} \right)} \right\}}}\left[ {{{\left\| {\hat \ell (w_t^n) - \ell (w_t^n)} \right\|}^2}} \right] \\
& = \sum\limits_{k = 1}^K {{d^k}} {\left( {w_t^n} \right)^2}\left( {\frac{1}{{{p^k}\left( {w_t^n} \right)}} - 1} \right)
\end{aligned}
\end{equation}

We formulate an optimization problem to minimize the variance. The reason why we try to minimize the variance is that the compressed parameters are closer to the original parameters when the variance decreases. Thus, the problem of minimizing the variance can be given as:

\begin{equation}
\begin{aligned}
\min \sum\limits_{k = 1}^K {\frac{{{d ^k}{{\left( {w_t^n} \right)}^2}}}{{{p^k}\left( {w_t^n} \right)}}} {\rm{ }}{\kern 1pt} {\kern 1pt} {\kern 1pt} {\rm{s}}{\rm{.t}}{\rm{.}}{\kern 1pt} {\kern 1pt} 0 < {p^k}\left( {w_t^n} \right) \le 1{\kern 1pt} {\kern 1pt} {\kern 1pt} {\rm{and}}{\kern 1pt} {\kern 1pt} \sum\limits_{k = 1}^K {{p^k}} \left( {w_t^n} \right) = {\varepsilon _t}
\end{aligned}
\end{equation}

Before solving the optimization problem, we first provide the following assumption of ${\varepsilon _t}$-balancedness:

\textbf{Assumption 4}. \emph{ $\ell \left( {w_{\rm{t}}^n} \right) = \sum\limits_{k = 1}^K {{d ^k}} \left( {w_{\rm{t}}^n} \right){\alpha ^k}\left( {w_{\rm{t}}^n} \right)$ is ${\varepsilon _t}$-unbalanced if ${d^k}\left( {w_{\rm{t}}^n} \right){\varepsilon _t} > {\kern 1pt} ||dw_t^n|{|_1}$. ${\varepsilon _t}$-balanced corresponds to the case that no element of $\ell \left( {w_{\rm{t}}^n} \right)$ is ${\varepsilon _t}$-unbalanced}.

Thus, the theorem for the solution of optimization problem can be given by:

\textbf{Theorem 2}. When $\ell \left( {w_{\rm{t}}^n} \right)$ is ${\varepsilon _t}$-balanced, solution for the aforementioned problem can be obtained by:

\begin{equation}
\begin{aligned}
{p^k}\left( {w_t^n} \right) = \frac{{{d ^k}\left( {w_t^n} \right){\varepsilon _t}}}{{{{\left\| {d \left( {w_t^n} \right)} \right\|}_1}}}
\end{aligned}
\end{equation}

\emph{Proof.} This can be proven by Lagrangian multiplier.

\textbf{Lemma 3} The different between compressed parameters ${\widehat \ell \left( {w_t^n} \right)}$ and uncompressed parameter ${\ell \left( {w_t^n} \right)}$ can be given by:

\begin{equation}
\begin{aligned}
{E_{\left\{ {{e^k}\left( {w_t^n} \right)} \right\}}}\left[ {{{\left\| {\widehat \ell \left( {w_t^n} \right) - \ell \left( {w_t^n} \right)} \right\|}^2}} \right] = \frac{{\delta _{1,n}^t}}{{{\varepsilon _t}}} + \delta _{2,n}^t
\end{aligned}
\end{equation}
where $\delta _{1,n}^t = \sum\limits_{k = 1}^K {\left( {{d^k}\left( {w_t^n} \right){{\left\| {d\left( {w_t^n} \right)} \right\|}_1}} \right)}$ and $\delta _{2,n}^t =  - \sum\limits_{k = 1}^K {{d^k}} {\left( {w_t^n} \right)^2}$.

\emph{Proof.} Based on Lemma 1 and Theorem 2, the ${E_{\left\{ {{e^k}\left( {w_t^n} \right)} \right\}}}\left[ {{{\left\| {\widehat \ell \left( {w_t^n} \right) - \ell \left( {w_t^n} \right)} \right\|}^2}} \right]$ can be denoted by:

\begin{equation}
\begin{aligned}
& {{E_{\left\{ {{e^k}\left( {w_t^n} \right)} \right\}}}\left[ {{{\left\| {\widehat \ell \left( {w_t^n} \right) - \ell \left( {w_t^n} \right)} \right\|}^2}} \right]}\\
& { = \sum\limits_{k = 1}^K {{d^k}} {{\left( {w_t^n} \right)}^2}\left( {\frac{1}{{{p^k}\left( {w_t^n} \right)}} - 1} \right)}\\
& { = \sum\limits_{k = 1}^K {{d^k}} {{\left( {w_t^n} \right)}^2}\left( {\frac{{{{\left\| {d\left( {w_t^n} \right)} \right\|}_1}}}{{{d^k}\left( {w_t^n} \right){\varepsilon _t}}} - 1} \right)}\\
& { = \frac{1}{{{\varepsilon _t}}}\sum\limits_{k = 1}^K {\left( {{d^k}\left( {w_t^n} \right){{\left\| {d\left( {w_t^n} \right)} \right\|}_1}} \right)}  - \sum\limits_{k = 1}^K {{d^k}} {{\left( {w_t^n} \right)}^2}}\\
& { = \frac{{\delta _{1,n}^t}}{{{\varepsilon _t}}} + \delta _{2,n}^t}
\end{aligned}
\end{equation}
where $\delta _{1,n}^t = \sum\limits_{k = 1}^K {\left( {{d^k}\left( {w_t^n} \right){{\left\| {d\left( {w_t^n} \right)} \right\|}_1}} \right)}$ and $\delta _{2,n}^t =  - \sum\limits_{k = 1}^K {{d^k}} {\left( {w_t^n} \right)^2}$.

After deriving the probability of variance of estimator, we are able to access the effects of compression on variance $\delta$ and communication time $T_{comm}$. Assuming that the model parameters to be uploaded is denoted as $\Upsilon $. Rather than sending original parameters, the devices can upload the compressed approximated parameters denoted as $\widehat \Upsilon $ with sparsity representation, which only uploads the ${\varepsilon _t}$ number of basic components. Thus, the communication time $T_{comm}$ can be rewritten as ${T_{comm}} = \gamma {\varepsilon _t}$, where $\gamma $ corresponds to the communication time of each device. As to variance $\delta $, the theorem for the variance of compressed locally aggregated parameters is based on the Theorem 2, which can be given by:

\textbf{Theorem 3}. The bound for the variance of compressed locally aggregated parameters is as follows:

\begin{equation}
\begin{aligned}
 {E_{\left\{ {\varphi _{t,i}^n} \right\},\left\{ {{e^k}\left( {w_t^n} \right)} \right\}}}& \left[ {{{\left\| {\hat \ell \left( {{w_t}} \right) - \nabla F\left( {{w_t}} \right)} \right\|}^2}} \right] \\
& \le \lambda {\left\| {\nabla F\left( {{w_t}} \right)} \right\|^2} + \frac{{{\delta _1}}}{{{\varepsilon _t}}} + {\delta _2}
\end{aligned}
\end{equation}
where $\lambda$, ${{\delta _1}}$ and ${\delta _2}$  are constants inversely proportional to mini-batch size.

\emph{Proof.} The ${E_{\left\{ {\varphi _{t,i}^n} \right\},\left\{ {{e^k}\left( {w_t^n} \right)} \right\}}}\left[ {{{\left\| {\hat \ell \left( {{w_t}} \right) - \nabla F\left( {{w_t}} \right)} \right\|}^2}} \right]$ can be described as:

\begin{equation}
\begin{aligned}
& {E_{\left\{ {\varphi _{t,i}^n} \right\},\left\{ {{e^k}\left( {w_t^n} \right)} \right\}}}\left[ {{{\left\| {\hat \ell \left( {{w_t}} \right) - \nabla F\left( {{w_t}} \right)} \right\|}^2}} \right] \\
& = {E_{_{\left\{ {\varphi _{t,i}^n} \right\},\left\{ {{e^k}\left( {w_t^n} \right)} \right\}}}}\left[ {||\hat \ell \left( {{w_t}} \right) - \ell \left( {{w_t}} \right)} \right.\left. { + \ell \left( {{w_t}} \right) - \nabla F\left( {{w_t}} \right)|{|^2}} \right] \\
&  \le {E_{\left\{ {{e^k}\left( {w_t^n} \right)} \right\}}}\left[ {||\hat \ell \left( {{w_t}} \right) - \ell \left( {{w_t}} \right)|{|^2}} \right] \\
& + {E_{\left\{ {\varphi _{t,i}^n} \right\}}}\left[ {{{\left\| {\ell \left( {{w_t}} \right) - \nabla F\left( {{w_t}} \right)} \right\|}^2}} \right]
\end{aligned}
\end{equation}

For the first term in Eqn. 19, the ${E_{\left\{ {{e^k}\left( {w_t^n} \right)} \right\}}}\left[ {||\hat \ell \left( {{w_t}} \right) - \ell \left( {{w_t}} \right)|{|^2}} \right]$ can be obtained by:

\begin{equation}
\begin{aligned}
& {E_{\left\{ {{e^k}\left( {w_t^n} \right)} \right\}}}\left[ {||\hat \ell \left( {{w_t}} \right) - \ell \left( {{w_t}} \right)|{|^2}} \right] \\
& \le \frac{1}{N}\sum\limits_{n = 1}^N {\sum\limits_{k = 1}^K {{d^k}} } {\left( {w_t^n} \right)^2}\left( {\frac{1}{{{p^k}\left( {w_t^n} \right)}} - 1} \right)\\
& = \frac{1}{N}\left( {\sum\limits_{n = 1}^N {\frac{{{\delta _{1,n}}}}{{{\varepsilon _t}}}}  + {\delta _{2,n}}} \right)
\end{aligned}
\end{equation}

where ${E_{\left\{ {{e^k}\left( {w_t^n} \right)} \right\}}}\left[ {{{\left\| {\hat \ell \left( {w_t^n} \right) - \ell \left( {w_t^n} \right)} \right\|}^2}} \right] = \sum\limits_{k = 1}^K {{d^k}} {\left( {w_t^n} \right)^2}\left( {\frac{1}{{{p^k}\left( {w_t^n} \right)}} - 1} \right)$ from Lemma 1, and $\sum\limits_{k = 1}^K {{d^k}} {\left( {w_t^n} \right)^2}\left( {\frac{1}{{{p^k}\left( {w_t^n} \right)}} - 1} \right) = \frac{{{\delta _{1,n}}}}{{{\varepsilon _t}}} + {\delta _{2,n}}$ from Lemma 3.

For the second term in Eqn. 19, ${E_{\left\{ {\varphi _{t,i}^n} \right\}}}\left[ {{{\left\| {\ell \left( {{w_t}} \right) - \nabla F\left( {{w_t}} \right)} \right\|}^2}} \right]$ can be obtained by $\lambda {\left\| {\nabla F\left( {{w_t}} \right)} \right\|^2} + \delta$ following by \cite{8th}.

Thus, based on Eqn. 19 and 20, the ${E_{\left\{ {\varphi _{t,i}^n} \right\},\left\{ {{e^k}\left( {w_t^n} \right)} \right\}}}\left[ {{{\left\| {\hat \ell \left( {{w_t}} \right) - \nabla F\left( {{w_t}} \right)} \right\|}^2}} \right]$ can be described by:

\begin{equation}
\begin{aligned}
 {E_{\left\{ {\varphi _{t,i}^n} \right\},\left\{ {{e^k}\left( {w_t^n} \right)} \right\}}}& \left[ {{{\left\| {\hat \ell \left( {{w_t}} \right) - \nabla F\left( {{w_t}} \right)} \right\|}^2}} \right] \\
& \le \lambda {\left\| {\nabla F\left( {{w_t}} \right)} \right\|^2} + \frac{{{\delta _1}}}{{{\varepsilon _t}}} + {\delta _2}
\end{aligned}
\end{equation}

where ${\delta _1} = {\max _k}\left( {\frac{1}{N}\sum\limits_{n = 1}^N {\delta _{1,n}^k} } \right)$ and ${\delta _2} = {\max _k}\left( {\frac{1}{N}\sum\limits_{n = 1}^N {\delta _{2,j}^k} } \right) + \delta $.

With the derived new communication time and variance, the Assumption 2 in Theorem 1 can be rewritten via Eqn. 22 where the new $\frac{{{\delta _1}}}{{{\varepsilon _t}}} + {\delta _2}$. To derive the error upper bound with parameter compression, we adapt Theorem 1 via new Assumption 2 and communication time, which considers the effects of parameter compression on communication time and variance. The updated Theorem is as follows:

\textbf{Theorem 4}. The bound error upper bound with parameter compression is as follows:

\begin{equation}
\begin{aligned}
{\upsilon _t}\left( {{I_t},{\varepsilon _t}} \right) & = \frac{{2\left[ {F\left( {{w_t}} \right) - {F_{{\rm{inf}}}}} \right]}}{{\eta T}}\left( {{T_{comp}} + \frac{{\gamma {\varepsilon _t}}}{{{I_t}}}} \right) \\ & + \frac{{\eta L(\frac{{{\delta _1}}}{{{\varepsilon _t}}} + {\delta _2})}}{N} + {\eta ^2}{L^2}(\frac{{{\delta _1}}}{{{\varepsilon _t}}} + {\delta _2})\left( {{I_t} - 1} \right)
\end{aligned}
\end{equation}

where ${\upsilon _t}\left( {{I_t},{\varepsilon _t}} \right)$ corresponds to the error upper bound considering local update and parameter compression.

\emph{Proof.} See Appendix in \cite{8th} except the new communication time ${{T_{comp}}}$ replaced by ${\gamma {\varepsilon _t}}$ and variance $\delta $ replaced by $\frac{{{\delta _1}}}{{{\varepsilon _t}}} + {\delta _2}$.

${\upsilon _t}\left( {{I_t},{\varepsilon _t}} \right)$ indicates the dynamics of trade-off between computation and communication, and trade-off between communication and precision, which is determined by ${{I_t}}$ and ${{\varepsilon _t}}$. The first term in ${\upsilon _t}\left( {{I_t},{\varepsilon _t}} \right)$ expression shows that the ${\upsilon _t}\left( {{I_t},{\varepsilon _t}} \right)$ decreases as ${{I_t}}$ increases because ${{I_t}}$ is in the denominator and the ${\upsilon _t}\left( {{I_t},{\varepsilon _t}} \right)$ decreases as ${{\varepsilon _t}}$ decreases due to ${{\varepsilon _t}}$ is in the numerator. Thus, the ${{I_t}}$ and ${{\varepsilon _t}}$ need to be reduce based on the analysis of the first term in Eqn. 22. However, the third term of Eqn. 22 require ${{I_t}}$ remain small due to ${{I_t}}$ is in the numerator. The second and the third term of Eqn. 22 require ${{\varepsilon _t}}$ remain large due to ${{\varepsilon _t}}$ is in the denominator. The above analysis indicates that either ${I_t} = 1$ or ${I_t} \gg 1$ is not an optimal choice due to the former results in unnecessary communication overheads, while the latter suffers from a prolonged convergence due to large discrepancy among local models caused by less communication. Either ${\varepsilon _t} = 1$ and ${\varepsilon _t} \gg 1$ is also not an optimal choice because the former sends imprecise model parameters causing the prolonged convergence, whereas the latter results in frequent communication. Thus, this paper aims at finding the optimal balance to adjust trade-offs between communication and computation/precision.

\section{The proposed EAFO algorithm}
The above theoretical analyses illustrate that the error upper bound is ruled by the ${I_t}$ and ${\varepsilon _t}$. This paper proposes efficient adaptive federated optimization (EAFO) algorithm which minimizes the learning error via jointly considering two variables including multiple local update and parameter compression and adaptively adjusting and balancing trade-offs between communication and computation/precision. Fig.1 presents an overview of EAFO scheme. Mathematically, EAFO finds the optimal balance to minimize the error upper bound in Eqn. 22, which can be denoted as:

\begin{equation}
\begin{aligned}
I_t^*,\varepsilon _t^* = \arg {\min _{I_t,{\varepsilon _t}}}{\upsilon _t}\left( {{I_t},{\varepsilon _t}} \right)
\end{aligned}
\end{equation}

This paper presents Theorem 5 to provide the theoretical analysis for the proof of convexity of ${\upsilon _t}\left( {{I_t},{\varepsilon _t}} \right)$, following by the Theorem 6 that finds the optimal solutions to solve the proposed problem of Eqn. 23.

\textbf{Theorem 5}. Let L, T and $\eta$ be defined therein. Choose assumptions (i) ${I_t} \ge 2$, (ii) ${\eta ^5} \approx 0$, (iii) $\left( {{L^4}T{\delta _1}/2\alpha \left( {F\left( {{w_t}} \right) - {F_{{\rm{inf}}}}} \right)\varepsilon _t^4} \right) < \infty $, (iv) $2{\eta ^2}LT{\delta _1}{I_t} \ge \alpha N\varepsilon _t^2\left( {F\left( {{w_t}} \right) - {F_{{\rm{inf}}}}} \right)$, Thus the ${\upsilon _t}\left( {{I_t},{\varepsilon _t}} \right)$ is convex.

\emph{Proof.}  Hessian matrix of ${\upsilon _t}\left( {{I_t},{\varepsilon _t}} \right)$  must be positive semidefinite is the condition of ${\upsilon _t}\left( {{I_t},{\varepsilon _t}} \right)$ to be convex. This paper derives the Hessian matrix of ${\upsilon _t}\left( {{I_t},{\varepsilon _t}} \right)$ as the followings：

\begin{equation}
\begin{aligned}
H\left( {{\upsilon _t}\left( {{I_t},{\varepsilon _t}} \right)} \right) = \left[ {\begin{array}{*{20}{c}}
{2X\frac{{\alpha {\varepsilon _t}}}{{I_t^3}}}&{ - X\frac{\alpha }{{I_t^2}} - P\frac{{{\delta _1}}}{{\varepsilon _t^2}}}\\
{ - X\frac{\alpha }{{I_t^2}} - P\frac{{{\delta _1}}}{{\varepsilon _t^2}}}&{2Z\frac{{{\delta _1}}}{{\varepsilon _t^3}} + 2P\left( {{I_t} - 1} \right)\frac{{{\delta _1}}}{{\varepsilon _t^3}}}
\end{array}} \right]
\end{aligned}
\end{equation}

where $X = \frac{{2\left[ {F\left( {{w_0}} \right) - {F_{{\rm{inf}}}}} \right]}}{{\eta T}},Z = \frac{{\eta L}}{N},{\rm{ and}}{\kern 1pt} {\kern 1pt} {\kern 1pt} P = {\eta ^2}{L^2}$. The diagonal elements and the determinant are positive is the condition of the Hessian matrix to be positive semidefinite, which can provide the proof of the convexity of ${\upsilon _t}\left( {{I_t},{\varepsilon _t}} \right)$. Obviously, the elements on the diagonal are positive. And the determinant can be described as the followings based on assumptions (i), (ii), (iii), and (iv).

\begin{equation}
\begin{aligned}
& {4\alpha {\delta _1}\left( {\frac{{{\rm X}Z}}{{I_t^3\varepsilon _t^2}} + \frac{{XP\left( {{I_t} - 1} \right)}}{{I_t^3\varepsilon _t^2}} - \left( {\frac{{{X^2}\alpha }}{{{\delta _1}I_t^4}} + \frac{{{P^2}{\delta _1}}}{{\alpha \varepsilon _t^4}} + \frac{{2XP}}{{I_t^2\varepsilon _t^2}}} \right)} \right)}\\
& { = 4\alpha {\delta _1}\left( {\frac{{4XZ}}{{I_t^3\varepsilon _t^2}} + \frac{{2XP}}{{I_t^2\varepsilon _t^2}} - \left( {\frac{{{X^2}\alpha }}{{{\delta _1}I_t^4}} + \frac{{{P^2}{\delta _1}}}{{\alpha \varepsilon _t^4}} + \frac{{4XP}}{{I_t^3\varepsilon _t^2}}} \right)} \right)}\\
& { = \frac{{2XP}}{{I_t^2\varepsilon _t^2}}\left( {1 - \frac{2}{{{I_t}}}} \right) + \frac{{4XZ}}{{I_t^3\varepsilon _t^2}} - \left( {\frac{{{X^2}\alpha }}{{{\delta _1}I_t^4}} + \frac{{{P^2}{\delta _1}}}{{\alpha s_k^4}}} \right)}\\
& { \ge \frac{{4Z}}{{I_t^3s_t^2}} - \left( {\frac{{X\alpha }}{{{\delta _1}I_k^4}} + \frac{{{P^2}{\delta _1}}}{{X\alpha \varepsilon _t^4}}} \right)}\\
& { = \frac{{4Z}}{{I_t^3\varepsilon _t^2}} - \left( {\frac{{X\alpha }}{{{\delta _1}I_t^4}} + \frac{{{\eta ^5}{L^4}T{\delta _1}}}{{2\alpha \left[ {F\left( {{w_0}} \right) - {F_{{\rm{inf}}}}} \right]\varepsilon _t^4}}} \right)}\\
& { \approx \frac{{4Z}}{{I_t^3\varepsilon _t^2}} - \frac{{X\alpha }}{{{\delta _1}I_t^4}}}\\
& { = \frac{1}{{N\eta T{\delta _1}\varepsilon _t^2I_t^4}}\left( {2\eta LT{\delta _1}{I_t} - \alpha N\varepsilon _t^2\left( {F\left( {{w_0}} \right) - {F_{{\rm{inf}}}}} \right)} \right)}\\
& { \ge 0}
\end{aligned}
\end{equation}

\textbf{Theorem 6}. The optimal solutions to minimize the error upper bound ${\upsilon _t}\left( {{I_t},{\varepsilon _t}} \right)$ can be given by:

\begin{equation}
\begin{aligned}
{I_t} = \sqrt {\frac{{2\alpha \left[ {F\left( {{w_t}} \right) - {F_{{\rm{inf}}}}} \right]\varepsilon _t^2}}{{{\eta ^3}{L^2}\left( {{\delta _1} + {\delta _2}{\varepsilon _t}} \right)}}}
\end{aligned}
\end{equation}

\begin{equation}
\begin{aligned}
{\varepsilon _t} = \sqrt {\frac{{{\delta _1}{\eta ^2}LT\left( {1 - \eta L\left( {{I_t} - 1} \right)} \right){I_t}}}{{2\alpha \left[ {F\left( {{w_t}} \right) - {F_{{\rm{inf}}}}} \right]}}}
\end{aligned}
\end{equation}

With Assumptions 5 (${\delta _1} \ll {\delta _2}{\varepsilon _t}$), Assumptions 6 ($\eta L\left( {{I_t} - 1} \right) \ll 1$) and Assumptions 7 (${F_{\inf }} = 0$), the ${I_t}$ and $\varepsilon _t$ can be approximated by:

\begin{equation}
\begin{aligned}
\frac{{{I_{t + 1}}}}{{{I_t}}} = & \sqrt {\frac{{F\left( {{w_{t + 1}}} \right) - {F_{{\rm{inf}}}}}}{{F\left( {{w_t}} \right) - {F_{{\rm{inf}}}}}}} \sqrt {\frac{{{\delta _1} + {\delta _2}{\varepsilon _t}}}{{{\delta _1} + {\delta _2}{\varepsilon _{t + 1}}}}} \frac{{{\varepsilon _{t + 1}}}}{{{\varepsilon _t}}} \\ \approx & \sqrt {\frac{{F\left( {{w_{t + 1}}} \right)}}{{F\left( {{w_t}} \right)}}} \sqrt {\frac{{{\varepsilon _{t + 1}}}}{{{\varepsilon _t}}}}
\end{aligned}
\end{equation}

\begin{equation}
\begin{aligned}
\frac{{{\varepsilon _{t + 1}}}}{{{\varepsilon _t}}} = & \sqrt {\frac{{F\left( {{w_t}} \right) - {F_{{\rm{inf}}}}}}{{F\left( {{w_{t + 1}}} \right) - {F_{{\rm{inf}}}}}}} \sqrt {\frac{{1 - \eta L\left( {{I_{t + 1}} - 1} \right)}}{{1 - \eta L\left( {{I_t} - 1} \right)}}} \sqrt {\frac{{{I_{t + 1}}}}{{{I_t}}}} \\ \approx & \sqrt {\frac{{F\left( {{w_t}} \right)}}{{F\left( {{w_{t + 1}}} \right)}}} \sqrt {\frac{{{I_{t + 1}}}}{{{I_t}}}}
\end{aligned}
\end{equation}

\emph{Proof.}  This can be proven via adopting the assumption 4-5 and setting the partial derivatives of ${\upsilon _t}({I_t},{\varepsilon _t})$ as 0.

Assumptions 5 is reasonable because ${\delta _2}$ is usually larger than ${\delta _1}$ and ${\varepsilon _t} \ge 1$. To reveal the reason why ${\delta _2} \gg {\delta _1}$, we first need to understand the source of ${\delta _1}$ and ${\delta _2}$. There are two sources of variance in federated learning when adopting both local update (SGD) and parameter compression. Performing parameter compression results in the first variance ${\delta _1}$ \cite{29th}. Adopting the mini-batch SGD rather than FGD accounts for the second variance ${\delta _2}$. The parameter compression will not distort the original sparse parameters due to the parameters are a sparse matrix which can be precisely reconstructed with few basic matrices \cite{24th}. Thus the first variance ${\delta _1}$ introduced by parameter compression is negligible compared with the second variance ${\delta _2}$. Assumptions 5 is reasonable due to the learning rate is small with the value of near 0.001 in practice, $({I_t} - 1) < 50$ and L < 1. Assumptions 7 is justifiable because ${F_{\inf }}$ is near zero \cite{30th}.

In Eqn. 28 and Eqn. 29, the values of ${I_t}$ and ${\varepsilon _t}$ are interdependent, which can be decoupled by substituting Eqn. 28 in Eqn. 29 as follows:

\begin{equation}
\begin{aligned}
\frac{{{I_{t + 1}}}}{{{I_t}}} = \sqrt[3]{{\frac{{F\left( {{w_{t + 1}}} \right)}}{{F\left( {{w_t}} \right)}}}}
\end{aligned}
\end{equation}

\begin{equation}
\begin{aligned}
\frac{{{\varepsilon _{t + 1}}}}{{{\varepsilon _t}}} = \sqrt[3]{{\frac{{F\left( {{w_t}} \right)}}{{F\left( {{w_{t + 1}}} \right)}}}}
\end{aligned}
\end{equation}

With the initial values ${I_0},{\varepsilon _0},F({w_0})$, the Eqn. 30-31 can be rewritten as:

\begin{equation}
\begin{aligned}
{I_t} = \sqrt[3]{{\frac{{F\left( {{w_t}} \right)}}{{F\left( {{w_0}} \right)}}}}{I_0}
\end{aligned}
\end{equation}

\begin{equation}
\begin{aligned}
{\varepsilon _t} = \sqrt[3]{{\frac{{F\left( {{w_0}} \right)}}{{F\left( {{w_t}} \right)}}}}{\varepsilon _0}
\end{aligned}
\end{equation}

Equation. 32-33 illustrates that as the loss value ${F\left( {{w_t}} \right)}$ decreases during training process of federated learning, ${I_t}$ needs to decrease and ${\varepsilon _t}$ should increase.

Algorithm 1 describes the details of the training process when performing EAFO algorithm. Full flow of Algorithm 1 can be described as the following steps: (1) the server broadcasts the calculated local update coefficient ${I_t}$, compression coefficient ${\varepsilon _t}$ and latest weights to the selected devices. The coefficient of local update determines the number of parameter computations in local training, whereas the coefficient of parameter compression determines the rate of parameter compression. (2) The devices perform multiple local training based on received local update coefficient ${I_t}$ and upload the compressed locally aggregated model parameters at the certain ratio of compression determined by compression coefficient ${\varepsilon _t}$. (3) The server aggregates all received compressed model parameters to update the global model. (4) The server jointly optimizes and adjusts the two variables including local update coefficient and parameter compression coefficient based on the latest value of loss via the Theorem 6. The latest local update coefficient, parameter compression coefficient and latest weights are broadcast by the server to the selected devices for next iteration.

\begin{algorithm}[htbp]\small
\caption{Efficient Adaptive Federated Optimization}
\label{alg:RU}
\begin{algorithmic}[1]
\STATE \textbf{Server Executes}:
\STATE \textbf{the initialization of the global model parameter}: $w_0$
\FOR { $round$ $t$ $=$ $1$ $to$ $I$}
\STATE \text{The server jointly optimizes ${I_t}$ and ${\varepsilon _t}$ via (30); }
\STATE \text{The serve broadcasts ${I_t}$ and ${\varepsilon _t}$ to selected devices; }
\FOR { $device$ $n$ $=$ $1$ $to$ $N$}
\STATE \text{Devices receive ${I_t}$ and ${\varepsilon _t}$; }
\FOR { $local$ $training$ $i$ $=$ $1$ $to$ $I_t$}
\STATE \text{Devices compute $\ell \left( {w_{t,i}^n,\varphi _{t,i}^n} \right)$; }
\STATE \text{Devices update via ${w_{t,i}^n}$ Eqn. 2; }
\ENDFOR
\STATE \text{Devices compute $\ell \left( {w_t^n} \right)$ via (3); }
\STATE \text{Devices compress $\ell \left( {w_t^n} \right)$ via (9) and (14); }
\STATE \text{Devices upload compressed weights $\hat \ell \left( {{w_t}} \right)$ to the server; }
\ENDFOR
\STATE \text{The server aggregates compressed weights via (4); }
\STATE \text{The server update the global model via (5); }
\STATE \text{The latest weights ${w_{t + 1}}$ are broadcast to the devices; }
\ENDFOR
\end{algorithmic}
\end{algorithm}

\section{EXPERIMENTAL EVALUATION}
This section evaluates the performance of the proposed EAFO in speeding up federated learning and compares it with the-state-of-the-art on Fashion-MNIST and CIFAR-10 image classification tasks. In our experiment, this paper implements the algorithm via a centos7 server with Nvidia TITAN RTX GPU.


\subsection{Experiment Setup}
Training Datasets:  Following \cite{31th,32th}, this paper constructs both IID and Non-IID datasets from Fashion-MNIST and CIFAR-10. The Fashon-MNIST dataset consists of 10 categories images of fashion items including 70000 $28*28$ images. The CIFAR-10 dataset is comprised of 60000 examples with $32*32$ color pixels in 10 classes such as airplane, bird and racing car.

Implementation settings:  For settings of training, this paper initializes the iteration time as 200, local epoch as 1, batch size as 32 and learning rate as 0.01. The proposed design is examined on image classification tasks on Fashion-MNIST and CIFAR-10 datasets. For model architectures, this paper adopts the Convolutional Neural Network (CNN) with a fully connected layer, a softmax output layer and two convolutional layers with 32 and 64 channels respectively on Fashion-MNIST, and a light-weight ResNet18 on CIFAR-10.

Comparison baselines:  This paper compares the proposed EAFO with the two categories of the-state-of-the-art algorithms: (1) Multiple local updates that reduces the frequent transmissions of model parameters via performing local updates repeatedly rather than one local update. This paper considers ADACOMM \cite{16th} which dynamically adjusts local update coefficient with an adaptive strategies as baselines. (2) Parameter compression that reduces the amount of data to be transmitted via data compression techniques. We adopt ATOMO \cite{22th} which exploits the sparsity of parameters to compress parameters for comparisons.

Evaluation metrics:  This paper considers different dimensions to analyze the experiment results accurately: (1) Round completion time (efficiency metric). This paper defines the round completion time as the time needed to spent in one round of federated learning. The reason why we adopt round completion time instead of iteration /communication round is that unlike traditional machine learning where different iterations take nearly same time and thus iterations round reflects the convergence speed but it is inapplicable to FL \cite{33th,34th}. The key factors to determined the convergence speed of FL is round completion time and iteration round. The round completion time is various over time due to three time-varying delays components: (i) Download delays of global model parameters from the devices with different downlink speeds. (ii) Transmissions delays of compressed local model parameters from the devices with different compression ratios and uplink speeds. (iii) Computation delays of local model update from the devices with different local update coefficient and hardware constraints. (2) Global Learning accuracy (utility metric). This paper considers global learning accuracy as average accuracy of global model to evaluate effectiveness of the proposed design to speed up the federated learning.

\subsection{Results and Analysis}

This paper compares the proposed EAFO with two categories of the-state-of-the-art algorithms including ADACOMM \cite{16th} of category of Multiple local updates and ATOMO \cite{22th} of category of Parameter compression. The comparative experiment results in Figs. 4a, 4b, and 4c present the values of accuracy, ${I_t}$ and ${\varepsilon _t}$ over round completion time for 32 devices on Fashion-MNIST datasets. Fig. 4a illustrates the superiority of the proposed EAFO in terms of efficiency and high accuracy, which achieves higher accuracies faster over round completion time. The values of ATOMO is assigned as 1 because it does not adopt multiple local update in Fig. 4b. Fig. 4c shows the values of ${\varepsilon _t}$ except for ADACOMM due to it does not adopt parameter compression whereas ATAMO adopts parameter compression with a fixed values of ${\varepsilon _t}$. The proposed EAFO differs from these algorithms in that the EAFO dynamically adjusts the values of ${I_t}$ and ${\varepsilon _t}$ during federated learning. For ${I_t}$, the proposed EAFO begins with high values to enable perform more local update but ends with lower values and high accuracy. For ${\varepsilon _t}$, the proposed EAFO begins with low values due to coarse parameters can still contribute to improving the accuracy of the model effectively with low accuracy but ends with high values providing finer grained parameters and high accuracy. The uplink and downlink data rates are set equal to 100 Kbps. The dataset is in IID setting. Note that ${I_t} \in \{ 1, \cdots ,30\}$ and ${\varepsilon _t} \in [4,8]$.


This paper evaluates performance with different uplink/downlink data rates of the proposed EAFO compared with two algorithms namely ADACOMM and ATOMO. The results in Figs. 5a, 5b, and 5c present the accuracy for 10/10 Mbps, 10/100 Kbps, and 100/10 Kbps uplink/downlink data rates. This paper considers high uplink/downlink data rates as 10 Mbps in Fig. 5a, where parameter compression is less important than computation with the relatively high data rates. In this case, computation has becomes more of a bottleneck than communication for scaling up federated learning. Fig. 5a presents that the ADACOMM outperforms the ATOMO which adopts parameter compression in 10/10 Mbps data rates, and the proposed EAFO outperforms the ADACOMM because of benefit from the parameter compression. In Fig. 5b considering the 10/100 Kbps data rates where the uplink communication has becomes the bottleneck, the ATOMO win the race with ADACOMM because it compresses the parameters in the uplink. Additionally, the proposed EAFO outperforms the ATOMO due to benefit from the local update. On the contrary, in Fig. 5c considering the 100/10 Kbps data rates, the downlink communication has becomes the bottleneck. Thus, the performance of ATOMO is similar to the performance of ADACOMM despite parameter compression in the uplink. However, the proposed EAFO outperforms both ATOMO and ADACOMM due to the benefit not only from local update but also parameter compression.

This paper investigates non-IID data distribution over devices for the proposed EAFO in Fig. 7. The results in Figs. 7 demonstrate that for non-IID settings where the efficiency improvement becomes increasingly challenging, the proposed EAFO is still on top compared with both ATOMO and ADACOMM even if all algorithms exhibit minor deterioration of performance in non-IID setting. The proposed EAFO can still speed up federated learning with a jitter curve experiencing fast upward trend.


\section{CONCLUSION}
Recently, FL has received considerable attention in IoT because of the capability of enabling devices to train machine learning models collaboratively via sharing local model parameters rather than privacy-sensitive data. Although only model parameters are shared in FL, FL-IoT implementation still suffers from the inefficiency and limited convergence performance, which become the significant bottleneck for scaling up federated learning for IoT. This paper proposes a novel efficient adaptive federated optimization algorithm (EAFO) to speed up federated learning convergence via jointly balancing trade-offs among communication, computation and model precision. The proposed EAFO minimizes the learning error by the joint consideration of two variables including local update coefficient and parameter compression coefficient. Additionally, the EAFO enables federated learning to adaptively and dynamically adjust the trade-offs among communication, computation and model precision. The evaluation results demonstrate that the superiority of the EAFO in terms of efficiency and accuracy compared with state-of-the-art algorithms. The proposed algorithm is complementary to existing approaches to accelerate federated learning convergence, which is more effective to further enhance efficiency of federated learning.

\section{ACKNOWLEDGMENTS}
This work was supported by the National Natural Science Foundation of China $\left( 62171049 \right)$ and the National Key R and D Program of China $\left(2020YFB1807805 \right)$.

\ifCLASSOPTIONcaptionsoff
  \newpage
\fi

\bibliography{Ref}

\end{document}